\documentclass{llncs}

\usepackage{makeidx}  
\usepackage{etoolbox} 
\makeatletter
\patchcmd{\ps@headings}{\rlap{\thepage}}{}{}{}
\patchcmd{\ps@headings}{\llap{\thepage}}{}{}{}
\makeatother
\usepackage{amsmath}
\usepackage{amssymb}
\usepackage{amsfonts}

\usepackage{subcaption}

\usepackage{booktabs} 
\usepackage[pdftex]{graphicx}
\PassOptionsToPackage{dvipsnames}{xcolor}
\usepackage{tikz}
\usepackage{siunitx}
\usepackage{multirow}
\usepackage{bm}
\usepackage{bbm}
\usepackage{cases}
\usepackage[ruled,vlined]{algorithm2e}
\usepackage{algorithmicx}
\usepackage{algpseudocode}
\usepackage[maxfloats=256]{morefloats}
\usepackage{colortbl,array,xcolor}
\usepackage{mathtools}
\usepackage[whole]{bxcjkjatype}
\usepackage{url}
\usepackage{float}
\usepackage{hyperref}

\usepackage{mathtools}

\definecolor{customgreen}{RGB}{0,200,0}
\newcommand{\greencheck}{\textcolor{customgreen}
{\checkmark}}

\SetKwInput{KwInput}{Input}
\SetKwInput{KwOutput}{Output}
\SetKwInput{KwInit}{Init}

\usetikzlibrary{positioning}
\usetikzlibrary{arrows,shapes,automata,backgrounds,fit,petri,calc}

\tikzstyle{latent} = [circle,fill=white,draw=black,inner sep=1pt,
minimum size=25pt, font=\fontsize{10}{10}\selectfont, node distance=1]

\tikzstyle{pseud} = [minimum size=25pt, font=\fontsize{10}{10}\selectfont, node distance=1]

\colorlet{DPNColor}{orange}
\colorlet{RCANColor}{RubineRed}

\newcommand{\mA}{\mathbf{A}} 
\newcommand{\mI}{\mathbf{I}} 
\newcommand{\mR}{\mathbf{R}} 
\newcommand{\vt}{\mathbf{t}} 

\newcommand{\mRrod}{\mR_{\textrm{rod}}} 

\newcommand{\vp}{\mathbf{p}} 
\newcommand{\vq}{\mathbf{q}} 
\newcommand{\vn}{\mathbf{n}} 
\newcommand{\vc}{\mathbf{c}} 
\newcommand{\vb}{\mathbf{b}} 
\newcommand{\vv}{\mathbf{v}} 
\newcommand{\vx}{\mathbf{x}} 
\newcommand{\vy}{\mathbf{y}} 
\newcommand{\va}{\mathbf{a}} 
\newcommand{\vu}{\mathbf{u}} 

\newcommand{\vpp}{\mathbf{p}'} 
\newcommand{\vzero}{\mathbf{0}} 

\newcommand{\dd}{\delta d} 

\newcommand{\ddSigma}{\dd_{\Sigma}}

\newcommand{\mRSigma}{\mR_{\Sigma}}
\newcommand{\vtSigma}{\vt_{\Sigma}}

\newcommand{\vomega}{\bm{\omega}}

\newcommand{\Sf}{\mathcal{S}^f}
\newcommand{\So}{\mathcal{S}^o}

\newcommand{\pF}{\partial \mathcal{F}}
\newcommand{\pO}{\partial \mathcal{O}}

\newcommand{\T}{\mathcal{T}}
\newcommand{\HpO}{\mathcal{H}_{\pO}}
\newcommand{\W}{\mathcal{W}}

\begin{document}

\newcommand\todo[1]{\textcolor{red}{[TODO: #1]}}

%

%
\mainmatter              
\title{
    Disentangled Iterative Surface Fitting\\for Contact-stable Grasp Planning
}
\titlerunning{Hamiltonian Mechanics}  
%
\author{
    Tomoya Yamanokuchi\inst{1} 
    \and Alberto Bacchin\inst{2} 
    \and Emilio Olivastri\inst{2}
    \and \\ Takamitsu Matsubara\inst{1} 
    \and Emanuele Menegatti\inst{2} 
}
\authorrunning{
    Tomoya Yamanokuchi et al.
} 
%
\tocauthor{
    Tomoya Yamanokuchi, Alberto Bacchin, Emilio Olivastri, Takamitsu Matsubara, and Emanuele Menegatti
}
\institute{
    Division of Information Science, Graduate School of Science and Technology, \\Nara Institute of Science and Technology, Nara, Japan, \\
    \email{[yamanokuchi.tomoya.ys9, takam-m]@is.naist.jp}, \\
    \and Department of Information Engineering, University of Padua, Padua, Italy, \\
    \email{emilio.olivastri@phd.unipd.it}, \\
    \email{[bacchinalb, emg]@dei.unipd.it}
}

\maketitle              

\begin{abstract}
    In this work, we address the limitation of surface fitting-based grasp planning algorithm, which primarily focuses on geometric alignment between the gripper and object surface while overlooking the stability of contact point distribution, 
    often resulting in unstable grasps due to inadequate contact configurations.  
    To overcome this limitation, we propose a novel surface fitting algorithm that integrates contact stability while preserving geometric compatibility.  
    Inspired by human grasping behavior, our method disentangles the grasp pose optimization into three sequential steps: (1) rotation optimization to align contact normals, (2) translation refinement to improve Center of Mass (CoM) alignment, and (3) gripper aperture adjustment to optimize contact point distribution.
    We validate our approach through simulations on ten YCB dataset objects, demonstrating an $80$\% improvement in grasp success over conventional surface fitting methods that disregard contact stability.
    Further details can be found on our project page\footnote{project page: $\texttt{\textcolor{blue}{https://tomoya-yamanokuchi.github.io/disf-project-page/}}$}.  
    
\end{abstract}

    \begin{figure}[thbp]
        \centering
        \includegraphics[width=\columnwidth]{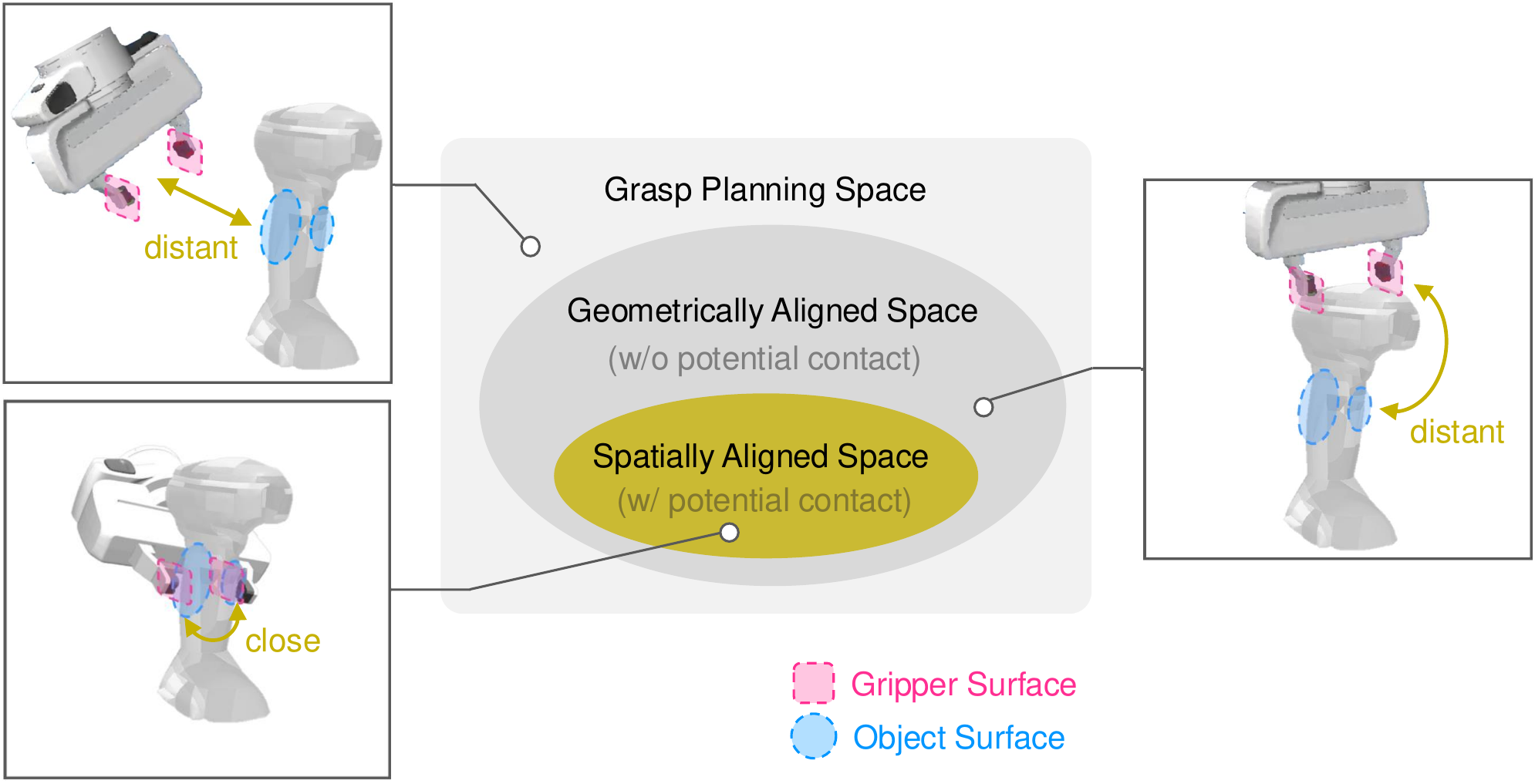}
        \caption{   
            The relationship between the grasp planning space, geometrically aligned space, and spatially aligned space.  
        }        \label{fig:relationship_of_contact_stablegrasp_space}
    \end{figure}

\section{Introduction}    
    Observations of human grasping behavior suggest that aligning the Center of Mass (CoM) of the hand closer with that of the object improves grasp stability~\cite{CoM1,CoM2,CoM3}.
    This is because CoM misalignment induces large rotational moment, which can destabilize the grasp~\cite{moment}.   
    Following this principle, numerous bio-inspired algorithms have been proposed to determine optimal grasp configuration~\cite{biomechanics_ellipsoid_1992,biomechanics_JoB_2005}. 
    However, while these algorithms achieve high accuracy in predicting human grasping behavior, they are fundamentally limited by the assumption that objects can be represented using simple geometric models, such as cylinders~\cite{biomechanics_EMBC_2011,biomechanics_IJIDeM_2012} or spheres~\cite{biomechanics_BioRob_2014}, making them unable to generalize to complex geometries.

    To overcome these limitations, grasp planning algorithms that do not rely on mathematical models of object shapes have been proposed in recent years, utilizing point cloud data~\cite{Fan_Case2018}.
    This approach builds upon the framework of 3D point cloud registration, which has been well established in the field of Computer Vision~\cite{ICP_TPAM_1987,ICP_point2plane_2001}. 
    By representing both the object and the robot hand's gripper surface as point cloud data, and directly optimizing their \textit{geometric compatibility} as an evaluation metric, this method determines an appropriate grasp pose.
    A series of studies by Fan et al. have demonstrated the effectiveness of this geometric compatibility-based optimization approach for grasp planning across a wide range of object shapes~\cite{Fan_IROS_2018,Fan_IROS_2019,Fan_RAL_2019,Fan_Sensors_2024}.

    While surface fitting algorithms based on geometric compatibility optimization offer high flexibility, they do not sufficiently account for whether the aligned surfaces actually lead to a stable grasp. 
    Specifically, achieving a stable grasp requires the ability to generate contact forces that can fully counteract external forces and torques (known as force-closure property~\cite{force_closure}). 
    However, by focusing solely on geometric alignment, these methods fail to consider fundamental factors necessary for generating contact forces, such as the appropriate spatial relationship between the hand and the object. 
    As a result, even if the surfaces are geometrically well-aligned, a spatial gap can form between the hand and the object, preventing actual contact from being established, or leading to an unstable distribution of contact points. 
    
    To address this issue, it is essential not only to align surfaces based on geometric compatibility but also to ensure that the hand and object surfaces are spatially well-aligned, allowing contact to be potentially established. 
    We refer to this spatial alignment, which facilitates contact, as \textit{contact stability} (Fig.~\ref{fig:relationship_of_contact_stablegrasp_space}).

    \begin{figure}[H]
        \centering
        \includegraphics[width=\columnwidth]{figures/disf_concept_v6-crop.pdf}
        \caption{   
            Overview of the proposed DISF optimization process. The grasp pose optimization is disentangled into three sequential steps: (1) rotation optimization to align contact normals, (2) translation refinement for Center of Mass (CoM) alignment, and (3) gripper aperture adjustment to optimize contact point distribution. Each step iteratively updates the gripper transformation parameters to ensure both geometric compatibility and contact stability. The arrows indicate the optimization flow, illustrating how the gripper adapts to the object surface through iterative surface fitting.
        }
        \label{fig:disf_algprithm}
    \end{figure}

    In this study, we propose a novel surface fitting-based grasp planning algorithm that incorporates contact stability alongside geometric compatibility, which we call Disentangled Iterative Surface Fitting (DISF). 
    From the perspective of contact stability, we explicitly integrate CoM alignment into the optimization process, drawing inspiration from the observation that, as mentioned earlier, humans naturally align their hand's CoM with that of the object to enhance grasp stability~\cite{CoM1,CoM2,CoM3}. 
    To achieve this, we leverage another key insight from human grasping behavior-that different pose parameters are optimized sequentially rather than simultaneously~\cite{human_grasp_analysis_2014}-and disentangle the overall grasp pose optimization into the following three sequential stages:
    (1) rotation optimization to align contact normals,  
    (2) translation refinement for CoM alignment, and  
    (3) gripper aperture adjustment to optimize contact point distribution.   
    Our disentangled optimization framework preserves the advantages of flexible geometric compatibility evaluation while systematically enhancing contact stability through CoM alignment. The overview of our DISF framework is shown in Fig.~\ref{fig:disf_algprithm}.

    To evaluate the effectiveness of our proposed method, we assessed the grasp quality planned by DISF based on two criteria: geometric compatibility, measured by geometric misalignment, and contact stability, measured by CoM misalignment.  
    Additionally, we evaluated the feasibility of the planned grasps in a physics simulator using ten objects from the YCB dataset.  
    Experimental results demonstrate that our method successfully reduces both geometric and CoM misalignment, thereby facilitating contact establishment between the hand and the object and ultimately achieving an $80$\% higher grasp success rate compared to conventional surface fitting methods that disregard contact stability.

\section{Related Works}
    \subsection{Effects of Center of Mass on Human Grasping Behavior}
        In human grasping, aligning the hand's CoM with the object's CoM enhances grasp stability by increasing contact area, stabilizing force distribution, and reducing rotational moments. Studies have shown that humans predict an object's CoM and adjust their contact points accordingly.
        For instance, Lukos et al. compared contact point selection in grasping when the CoM position was known versus unknown, showing that contact points are adapted based on a predictable CoM~\cite{CoM1}.
        Furthermore, Desanghere \& Marotta investigated how CoM estimation through visual information influences the selection of grasp positions, reporting that fixation locations are sensitive to CoM and affect grasp locations~\cite{CoM2}.
        
        These findings highlight the crucial role of CoM alignment in human grasping behavior. Motivated by this principle, our study explores how to systematically incorporate CoM alignment into computational grasp planning, particularly within our surface fitting-based approach.

    \subsection{Computational Modeling of CoM on Grasp Pose Optimization}
        While CoM alignment has been shown to contribute to grasp stability in human grasping behavior, how it is incorporated into computational models remains an open question.
        Biomechanical studies have demonstrated that CoM alignment is achieved as a result of optimizing joint positions and contact points~\cite{biomechanics_JoB_2005}.
        Specifically, as finger joint configurations adapt to the object surface, stable contact is ensured, and CoM alignment is implicitly achieved in the process.
        Moreover, computational biology has proposed grasp pose prediction models that explicitly consider CoM alignment, such as Klein et al. \cite{CoM3}, who demonstrated that variations in an object's CoM influence grasp location choices. These findings suggest that grasp poses predicted through CoM-aware optimization can explain many aspects of human grasping behavior and that CoM alignment is likely an essential factor in grasp planning. Building upon these insights, this study integrates CoM alignment into a surface fitting-based grasp planning algorithm, enabling it to account not only for geometric compatibility but also for contact stability.

    \subsection{Grasp Planning via Iterative Surface Fitting}
        Optimal grasping has been shown to result in a posture where the hand and the object's surface align~\cite{biomechanics_JoB_2005,taxonomy_TRO_1989,taxonomy_THMS_2016}.
        This alignment ensures that the robot hand establishes stable contact with the object by properly configuring the contact points.
        From this perspective, grasp planning can be interpreted as a surface fitting problem, where recent studies have applied point cloud processing techniques from Computer Vision to optimize the geometric compatibility between the object and the robot hand~\cite{Fan_Case2018}. 
        Unlike traditional grasp pose prediction models studied in biomechanics, this approach does not require the object's shape to be mathematically well-defined.
        As a result, it enables a more flexible grasp planning framework that is not constrained by predefined geometric models and can be applied to a wide variety of object shapes~\cite{Fan_IROS_2018,Fan_IROS_2019,Fan_RAL_2019}.
        However, existing surface fitting algorithms evaluate geometric compatibility alone without considering contact point configuration, which can result in unstable grasps where contact fails to occur or the distribution of contact points is imbalanced.
        
        In this study, we build upon flexible surface fitting algorithms and newly integrate CoM alignment to propose a grasp planning algorithm that ensures the proper establishment of contact points.
        This enables a grasp planning framework that considers both geometric compatibility and contact stability.


\section{Preliminaries}
    \subsection{Contact Surface Optimization}
\begin{figure}[thpb]
        \centering
        \includegraphics[width=.9\columnwidth]{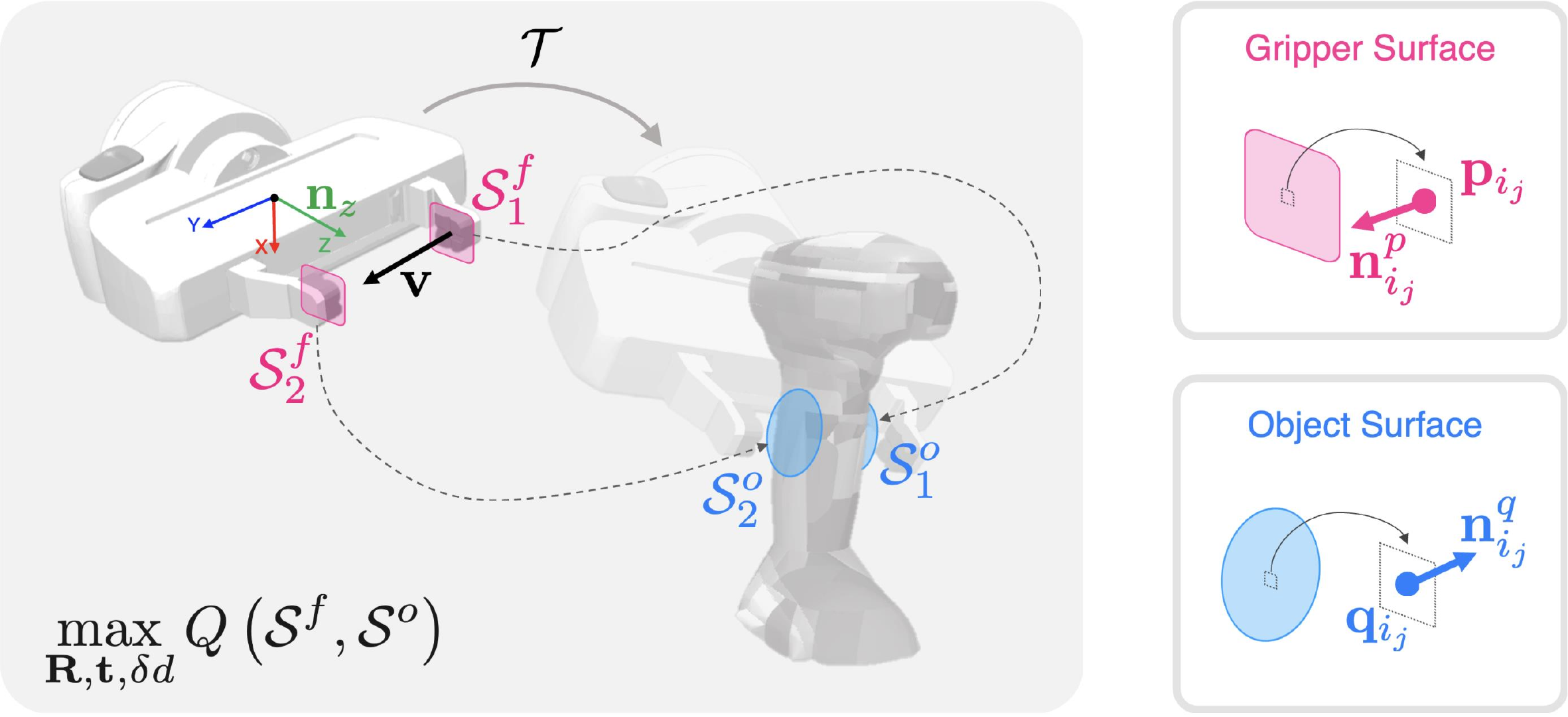}
        \caption{An image demonstrating how grasp planning can be reformulated as a contact surface optimization problem.}
        \label{fig:contact_surface_opt}
    \end{figure}

        The grasp planning problem with antipodal grippers can be modeled as a contact surface optimization problem which maximizes the grasp quality $Q$ by optimizing the rotation and translation parameter ($\mR, \vt$) as well as the fingertip displacement $\dd$ from the original gripper aperture $d$, given a specific set of contact surfaces between the fingertip and object:
        \begin{subequations}
        	\label{eq:general_form}
        	\begin{align}
        	\max_{\mR, \vt, \dd} &\  Q(\Sf, \So) \label{eq1:cost}\\
        	\mathrm{s.t.} \quad 
                & \Sf_j \subset \T(\pF_j, \mR, \vt, \dd), \quad j = 1,2 \label{eq1:surface_finger}\\
        	& \So_j = \HpO(\Sf_j), \quad j = 1,2 \label{eq1:surface_object}\\
        	& \Sf_j \in \W(d_0 + \delta d) \quad j = 1,2 \label{eq1:constraint1}\\
        	& d_0 + \dd \in [d_{\text{min}}, d_{\text{max}}] \label{eq1:constraint2}
        	\end{align}
        \end{subequations}        
        where $j \in \{1, 2\}$ is the finger index, $\Sf_j$ and $\So_j$ are the finger and object contact surfaces. 
        The $\Sf = [\Sf_1, \Sf_2]$ is the set of contact surfaces across the multiple fingers and the $\So = [\So_1, \So_2]$ is the corresponding set of contact surfaces for the object. 
        The finger contact surface lies on its canonical surface $\pF_j$ projected by the transformation function $\T$. 
        The object contact surface $\So_j$ is determined by a correspondence matching algorithm $\HpO$ given the object canonical surface $\pO$ and its query $\Sf_j$. 
        The contact surfaces are constrained by the gripper's working range \([d_{\text{min}}, d_{\text{max}}]\), defined by the robot's kinematics. The optimization aims to determine the optimal gripper transformation \((\mathbf{R}^*, \mathbf{t}^*, \delta d^*)\).
        This concept of contact surface optimization is illustrated in Fig.~\ref{fig:contact_surface_opt}. 

    \subsection{Iterative Surface Fitting}
        \subsubsection{Gripper Transformation}
            Given a specific contact point-normal pair $(\vp_{i_j}, \vn_{i_j}^p) \in \Sf_j$, where $i = 1, \dots, N$ and $N$ is the number of the pair for fingertip contact surfaces, the transformation function $\T$ for the gripper is defined as follows:
            \begin{equation}
                \label{eq:gripper_point_normal_set_transformation}
                \begin{aligned}
                    \T ((\vp_{i_j}, \vn_{i_j}^p), \mR(\vomega), \vt, \dd)  = (\mR(\vomega) \vp_{i_j} + \vt + 0.5(-1)^j {\mR(\vomega)}{\vv \dd}, \mR(\vomega) \vn_{i_j}^p)
                \end{aligned}
            \end{equation}
            where $\mR(\vomega)$ is the rotation matrix parameterized with the axis-angle vector $\vomega \in so(3)$, $\vv \in \mathbb{R}^3 $ is the unit vector pointing from $\Sf_1$ to $\Sf_2$. 

        \subsubsection{Grasp Quality Measures}
            \label{subsubsec:grasp_quality_measures}
            The geometric compatibility of point cloud data is generally evaluated using surface distance, called point-to-plane distance~\cite{ICP_point2plane_2001}.
            Therefore, grasp quality in surface fitting-based grasp planning is also assessed based on this criterion~\cite{Fan_Case2018,Fan_IROS_2018,Fan_IROS_2019}, which is defined as the distance between each point on the gripper surface and the tangent plane of the object surface:
            \begin{equation}
                \label{eq:Ep}
                \begin{aligned}
                     E_p(\vomega,\vt,\dd; \Sf, \So) &  = \sum_{i=1}^{N} \sum_{j = 1}^{2}\left((\vpp_{i_j}-\vq_{i_j})^\top \vn_{i_j}^q \right)^2, 
                \end{aligned}
            \end{equation}
            where $\vq_{i_j}$ and $\vn^q_{i_j}$ are the point and normal vector on the object contact surface $\So$. 
            
            In the context of grasp planning, normal misalignment is also an important criterion, as it directly affects the ability to achieve force-closure properties~\cite{force_closure}. 
            Stable contact requires the normals of the gripper and object surfaces to be oriented in opposite directions; thus, misalignment is measured by evaluating the deviation between these two normal vectors.
            \begin{equation}
                \label{eq:En}
                \begin{aligned}
                     E_n(\vomega; \Sf, \So)   = \sum_{i = 1}^N \left(\left(\mR(\vomega) \vn_{i}^p \right)^\top \vn_{i}^q + 1 \right)^2.
                \end{aligned}
            \end{equation}
            
            Additionally, approach direction misalignment~\cite{Fan_Sensors_2024}, is also often used. 
            In this study, we adopt this concept with modifications to better suit our formulation: 
            \begin{equation}
                \label{eq:Ea}
                \begin{aligned}
                     E_a(\vomega; \Sf, \So)   = \sum_{i = 1}^N \left(\left(\mR(\vomega) \vn_z \right)^\top \vn_{app} - 1\right)^2,
                \end{aligned}
            \end{equation}
            where $\vn_z$ is the z-axis direction of the gripper defined by the hand plane, and $\vn_{app}$ is the approach direction of the gripper. 
            This metric plays a crucial role in avoiding collisions and achieving a natural grasping posture.

        \subsubsection{Gradient-based Optimization with Iterative Least-squares Method}
            \label{sec:method:optimization}
            Direct optimization of the rotation matrix is challenging due to its constraint on the special orthogonal group $SO(3)$. 
            To address this, we approximate it as a small rotation around each axis~\cite{small_rotation}, allowing the rotation matrix to be expressed as:
            \begin{equation}
                \label{eq:R_skew}
                \mR(\vomega) \approx \mI + [\vomega]_{\times}
            \end{equation}
            where $\mI$ is $3 \times 3$ identity matrix, $[\cdot]_{\times}$ means skew-symmetric matrix, and $[\vomega]_{\times}$ is skew-symmetric matrix with the small rotation vector $\vomega = [\omega_x, \omega_y, \omega_z]^\top$: 
            \begin{equation}
                \label{eq3:closed_form}
                [\vomega]_{\times} = \left[\begin{array}{ccc}
                0         & \omega_z & \omega_y \\
                \omega_z  &  0       & -\omega_x \\
                -\omega_y & \omega_x & 0
                \end{array}\right].  
            \end{equation}
             This linearization simplifies the optimization process by reducing the complexity of handling the full rotation matrix $\mR$.  By applying the small rotation approximation to the grasp quality measures, the rotational component can be linearized into a form that is compatible with least-squares optimization, as follows:
            \begin{equation} 
            \label{eq:least_squares} 
                \min_{\vx} \|\mA \vx - \vb\|^2. 
            \end{equation}
            where $\vx$ contains the unknown parameters, e.g. $\vomega$, $\vt$, or $\dd$, $\mA$ is the coefficient matrix derived from the Jacobians of the grasp quality measures with respect to the parameters in $\vx$, and $\vb$ is the residual vector. 
            Taking the derivative of the squared error with respect to \( \vx \) and setting it to zero yields the normal equation $\mA^\top \mA \vx = \mA^\top \vb$ whose solution is given by $\vx = (\mA^\top \mA)^{-1} \mA^\top \vb$. 
            This equation is solved iteratively to maximize the grasp quality measures.

\section{Proposed Methods}
    \label{sec:method}
    Our goal is to integrate contact stability into the surface fitting algorithm.
    To achieve this, we take inspiration from human grasping behavior and introduce an optimization process for CoM alignment~\cite{CoM1,CoM2,CoM3}.
    To realize this approach, we attempt to disentangle and reconstruct conventional surface fitting algorithms.
    Specifically, inspired by observations that different posture parameters in human grasping are optimized sequentially~\cite{human_grasp_analysis_2014}, we disentangle the grasp pose optimization process into three stages instead of optimizing them simultaneously: 
    (1) rotation optimization to align contact normals, 
    (2) translation refinement for CoM alignment, and 
    (3) gripper aperture adjustment to optimize contact point distribution. 

    In this section, we first describe each of these disentangled optimization processes individually, followed by an explanation of the integrated Disentangled Iterative Surface Fitting (DISF) algorithm. 
        
    \subsection{Rotation Optimization for Contact Normal Misalignment (RO-CNM)}
            \begin{algorithm}[t]
                \caption{Rotation Optimization for Contact Normal Misalignment (RO-CNM)}
                \label{alg:PRO-CAM}
                \begin{algorithmic}[1] 
                    \State \textbf{Input:} $\pF, \Sf, \So, \mR^v, \vv$ 
                    \State $\vomega^* \gets \underset{\vomega}{\mathrm{min}} \, E_{na} \left(\vomega; \Sf, \So \right)$
                    \State $\mR^v \gets \mRrod(\vomega^*) \mR^v$
                    \State $\vv \gets \mR^v \vv$ 
                    \State $\Sf_j \gets \T(\Sf_j, \mRrod(\vomega^*), \vt=\vzero, \dd=0), \quad j=1,2$ 
                    \State $\pF_j \gets \T(\pF_j, \mRrod(\vomega^*), \vt=\vzero, \dd=0), \quad j=1,2$
                    \State \Return{$(\pF, \Sf, \vomega^*, \mR^v, \vv)$}
                \end{algorithmic}
            \end{algorithm}

            The first step is to optimize the rotation parameter $\vomega$ to align contact normals for the force-closure properties~\cite{force_closure}. 
            Therefore, it is considered independent of minimizing surface distance, which has been the focus of previous studies~\cite{Fan_Case2018,Fan_IROS_2018,Fan_IROS_2019}.
            At the same time, rotation also plays a role in aligning the approach direction.
            Based on this, we define the following grasp quality measure for optimizing the rotation parameters, which combines contact normal misalignment and approach direction misalignment with weighted contribution $\beta \in \mathbb{R}$:     
            \begin{equation}
                \label{eq:E_na}
                E_{na}(\vomega; \Sf, \So) = E_{n}(\vomega; \Sf, \So) + \beta^2 E_{a}(\vomega; \Sf, \So). 
            \end{equation}
            The overall optimization process for rotation parameters is summarized in Algorithm \ref{alg:PRO-CAM}.
            
            The palm rotation optimization can be formulated as a least-squares problem that is similar to Eq.~\eqref{eq:least_squares} with an augmented matrix $\Tilde{\mA} = [\mA_n^\top, \mA_a^\top]^\top$, an augmented residual $\Tilde{\vb} = [\vb_n, \vb_a]^\top$, and an unknown parameter $\vx = \vomega$, where $\mA_n = [\va_{n,1}^\top, \dots, \va_{n,N}^\top]^\top$ and $\vb_n = [b_{n,1}, \dots, b_{n,N}]^\top$ are parameters computed for the contact normal misalignment $E_n$, and $\mA_a = \va_{a}$ and $\vb_a = b_{a}^\top$ are parameters coming from the approach direction misalignment with: 
            \begin{subequations}
            	\label{eq:E_na_param}
            	\begin{align}
                	\va_{n,i} &= (\vn^p_i \times \vn^q_i)^\top, \label{eq:En_param:a}\\
                	b_{n,i} &= - \left( (\vn^p_i)^\top \vn^q_i  + 1 \right), \label{eq:En_param:b} \\
                	\va_{a} &= \beta (\vn_{z} \times \vn_{app})^\top, \label{eq:Ea_param:a}\\
                	b_{a} &= - \beta \left( (\vn_{z})^\top \vn_{app} - 1 \right). \label{eq:Ea_param:b}
                \end{align}
            \end{subequations} 

            Once the optimal parameter $\vomega^*$ is determined, it is converted into the optimal rotation matrix $\mRrod(\vomega^*)$ using the Rodrigues' rotation formula: 
            \begin{equation}
                \label{eq:R_rod}
                \begin{aligned}
                     \mRrod(\vomega) = \mI + \sin\theta [\vu]_{\times} + (1 - \cos\theta) [\vu]_{\times}^2, 
                \end{aligned}
            \end{equation}
            where $\mI$ is $3 \times 3$ identity matrix, $\vu = \vomega/\|\vomega\|$ describes rotation axis while $\theta = \|\vomega\|$ describes rotation angle. 
            The resulting rotation matrix is then used to update the current fingertip pointing vector $\vv$ (Line 3,4),  the fingertip contact surface $\Sf$ (Line 5), and its canonical surface $\pF$ (Line 6). 
            Finally, the necessary parameters are passed to the next step of translation refinement (Line 7).

        \subsection{Translation Refinement for CoM Alignment (TR-CoMA)}
            \begin{algorithm}[t]
                \caption{Translation Refinement for CoM Alignment (TR-CoMA)}
                \label{alg:PTR-CEN}
                \begin{algorithmic}[1] 
                    \State \textbf{Input:} $\pF, \pO, \Sf, \vv$
                    \State $\vc^p \gets \texttt{centroid}(\pF)$
                    \State $\vc^q \gets \texttt{centroid}(\pO)$
                    \State $\vt^c = \vc^q - \vc^p$
                    \State $\Sf_j \gets \T(\Sf_j, \mR=\mI, \vt=\vt^c, \dd=0), \quad j=1,2$
                    \State $\pF_j \gets \T(\pF_j, \mR=\mI, \vt=\vt^c, \dd=0), \quad j=1,2$
                    \State \Return{$(\pF, \Sf, \vt^c)$}
                \end{algorithmic}
            \end{algorithm}

            The second step is to refine the translation parameter $\vt$.  
            In this process, we perform a refinement to align the CoMs of the gripper and object surfaces.  
            Algorithm \ref{alg:PTR-CEN} outlines the entire procedure.  
            
            Since the true CoMs of both the gripper and the object cannot be directly computed from geometric data alone, we approximate them using the \textit{centroids} of their respective surfaces as surrogates.  
            To compute the centroid, we use the \texttt{centroid()} function, which calculates the centroid of an input surface $\partial$ as follows:  
            \begin{equation} 
                \label{eq:centroid}
                \begin{aligned}
                    \texttt{centroid}(\mathcal{\partial}) = \frac{1}{K}
                    \sum_{k=1}^K \vy_{k}, \quad \vy_k \in \partial, 
                \end{aligned}
            \end{equation}    
            where $K$ is the number of point cloud data representing the input surface (Line 2,3).              
            This refinement process utilizes the entire surface to align the CoMs, e.g., $\pF, \pO$, rather than the contact surface, in order to mitigate the effects of spatial bias in contact point selection.
            Finally, the obtained translation refinement parameter $\vt^c$ (Line 4) is used to update the current gripper surfaces, e.g., $\Sf$ and $\pF$ (Line 5,6), and the necessary parameters are passed to the next step of fingertip displacement optimization (Line 7).

        \subsection{Fingertip Displacement Optimization for Stable Contact Distribution (FDO-SCD)}
            Given the appropriate grasp pose ($\mR(\vomega^*), \vt^*$), the final step it to optimize the fingertip displacement $\dd$. 
            The overall procedures is shown in Algorithm \ref{alg:FDO-SCD}. 
            
            The goal of this optimization is to align the gripper surfaces with the object surface in terms of distance for stable distribution of contact points. 
            This is achieved by solving the following one-dimensional least-squares problem for $\dd$ to minimize the gripper-object surface distance defined by Eq.~\eqref{eq:Ep}: 
            \begin{equation}
                \min_{\delta d} \sum_{i=1}^N \sum_{j = 1}^{2} (b_{i_j} - a_{i_j}\delta d)^2, \quad 
                \text{s.t.} \quad \delta d + d_0 \in [d_\text{min}, d_\text{max}].
            \end{equation}
            where
            \begin{subequations}
            	\label{eq:Ep_param}
            	\begin{align}
                	a_{i_j} &= 0.5 (-1)^{j-1} (
                \vv)^\top \vn^q_{i_j}, \\
                	b_{i_j} &= \left(\vp_{i_j} - \vq_{i_j} \right)^\top \vn_{i_j}^q. 
                \end{align}
            \end{subequations} 
            Assuming a two-finger parallel gripper, the optimal fingertip relative motion is then given by 
            \begin{equation} 
                \label{eq:dd_solution}
                \delta d^* = \begin{cases} 
                d_\text{min} - d, & \text{if } \delta \hat{d} + d < d_\text{min}, \\
                \delta \hat{d}, & \text{if } d_\text{min} \leq \delta \hat{d} + d \leq d_\text{max} ,\\
                d_\text{max} - d, & \text{if } \delta \hat{d}  + d > d_\text{max}, \\
                \end{cases}
            \end{equation}
            with
            \begin{equation}
                \label{eq:dd_hat}
                \delta \hat{d} = \frac{\sum_{i=1}^m \sum_{j = 1}^{2}a_{i_j}b_{i_j}}{\sum_{i=1}^m \sum_{j = 1}^{2}a_{i_j}^2}.
            \end{equation}

            \begin{algorithm}[t]
                \caption{Fingertip Displacement Optimization for Stable Contact Distribution (FDO-SCD)}
                \label{alg:FDO-SCD}
                \begin{algorithmic}[1] 
                    \State \textbf{Input:} $\pF, \Sf, \So, \vv, d$
                    \State $\dd^* \gets \underset{\dd}{\mathrm{min}} \, E_p \left(\dd, d; \Sf, \So\right)$
                    \State $\Sf_j \gets \T(\Sf_j, \mR=\mI, \vt=\vzero, \dd^*), \quad j=1,2$
                    \State $\pF_j \gets \T(\pF_j, \mR=\mI, \vt=\vzero, \dd^*), \quad j=1,2$
                    \State \Return{$(\pF, \Sf, \dd^*)$}
                \end{algorithmic}
            \end{algorithm}

    \subsection{Disentangled Iterative Surface Fitting}
        This section presents the unified surface fitting algorithm.  
        The proposed Disentangled Iterative Surface Fitting (DISF) iteratively applies the three optimization steps to achieve feasible grasps with both geometric compatibility and contact stability. 
        The overall flow of DISF algorithm is shown in Algorithm \ref{alg:disf}. 
        
        The algorithm starts by initializing the rotation matrix, translation vector, and gripper aperture. 
        It then iteratively applies three optimization steps: (1) RO-CNM, (2) TR-CoMA, and (3) FDO-SCD. After each iteration, the updated parameters, including the cumulative rotation matrix \( \mRSigma \), translation vector \( \vtSigma \), and fingertip displacement \( \ddSigma \), are evaluated. 
        The process continues until the change in the surface updates, measured as $|e - e_p|$, falls below a predefined threshold $\Delta e$, at which point the algorithm terminates.

    \begin{algorithm}[t]
        \caption{
            DISF: Disentangled Iterative Surface Fitting
        }\label{alg:disf}
        \KwInput{
            $\pF$, 
            $\pO$, 
            $\mR_0$, 
            $\vt_0$, 
            $d_0$, 
            $d_{\textrm{min}}$, 
            $d_{\textrm{max}}$,
            $\vv_0$, 
            $\Delta e$
        }
        \KwInit{
            $\pF_0 \gets \T(\pF, \mR_0, \vt_0, d_0)$,
            $\mRSigma \gets \mI$, 
            $\vtSigma \gets \vzero$, 
            $\ddSigma = 0$,
            $d = d_0$, 
            $\vomega^* = \vzero$, 
            $\vt^* = \vzero$, 
            $\dd^* = 0$, 
            $e_p = \infty$, 
            $\vv = \vv_0$
        }
        \While{
            $e_p - e \leq \Delta e$
        }{
            $e_{p} \gets E\left(\Sf, \So\right)$\;
            $(\pF, \Sf, \vomega^*, \mR^v, \vv) \gets \texttt{RO-CNM}(\pF, \Sf, \So, \mR^v, \vv)$\;
            $(\pF, \Sf, \vt^*) \gets \texttt{TR-CoMA}(\pF, \pO, \Sf, \vv)$\;
            $(\pF, \Sf, \dd^*) \gets \texttt{FDO-SCD}(\pF, \Sf, \So, \vv, d)$\;
            $d \gets d + \delta d^*$\;
            $e \gets E\left(\Sf, \So\right)$\;
            $\mRSigma \gets \mRrod(\vomega^*) \mRSigma$\;
            $\vtSigma \gets \vt^* + \vtSigma$\;
            $\ddSigma \gets \dd^* + \ddSigma$\;
        }
        $\mR^* = \mRSigma \mR_0 $\;
        $\vt^* = \vtSigma + \vt_0$\;
        $\dd^* = \ddSigma $\;
        \Return{$(\mR^*, \vt^*, \dd^*)$}\;
    \end{algorithm}

\section{Simulation Experiments}
    \subsection{Grasp Quality Evaluation}
        \subsubsection{Settings}
            To validate the effectiveness of the proposed method, we evaluated the grasp quality for ten YCB objects~\cite{YCB_Dataset} using both the conventional geometric compatibility measures and the contact stability measure via CoM misalignment introduced in this study. 
            
            As an evaluation metric for geometric compatibility, we used the following weighted measure, which combines the surface distance defined in Eq.~\eqref{eq:Ep} and the contact normal misalignment defined in Eq.~\eqref{eq:En} with a scaling factor \( \alpha \): 
            \begin{equation} 
                \label{eq:E_geom}
                \begin{aligned}
                    E_{geom}(\vomega,\vt,\dd; \Sf, \So) = E_p(\vomega,\vt,\dd; \Sf, \So) + \alpha^2 E_{n}(\vomega; \Sf, \So). 
                \end{aligned}
            \end{equation} 
            The CoM misalignment was computed using the norm of the CoM difference between the gripper's and the object's canonical surfaces: 
            \begin{equation} 
                \label{eq:E_CoM}
                \begin{aligned}
                    E_{CoM}(\vomega,\vt,\dd; \Sf, \So) = \|\texttt{centroid}(\pO) - \texttt{centroid}(\pF^*) \|, 
                \end{aligned}
            \end{equation} 
            where $\pF^*$ is the canonical gripper surface transformed by the optimal grasping parameter $(\mR(\vomega^*), \vt^*, \dd^*)$. 

            As baseline methods, we used the conventional iterative surface fitting algorithm~\cite{Fan_Case2018} that does not consider CoM alignment (VISF: Vanilla Iterative Surface Fitting) and a sampling-based search algorithm using CMA-ES~\cite{CMA_ES}.

            In the experiments, we set the parameter $\alpha=0.1$, $\beta=0.8$, $d_0=0.091$, $d_{\mathrm{min}}=0.011$, $d_{\mathrm{max}}=0.091$, $\vv_0=[0, 1, 0]$, $\vn_{z0} = [0, 0, 1]$, $\Delta e = 1\mathrm{e}{-4}$. 
            We used the predefined approach direction $\vn_{app}$ for each object. 
            The more details of the experimental settings can be found in our \href{https://tomoya-yamanokuchi.github.io/disf-project-page/}{project page}.

            \begin{figure}[thbp]
                \centering
                \includegraphics[width=\columnwidth]{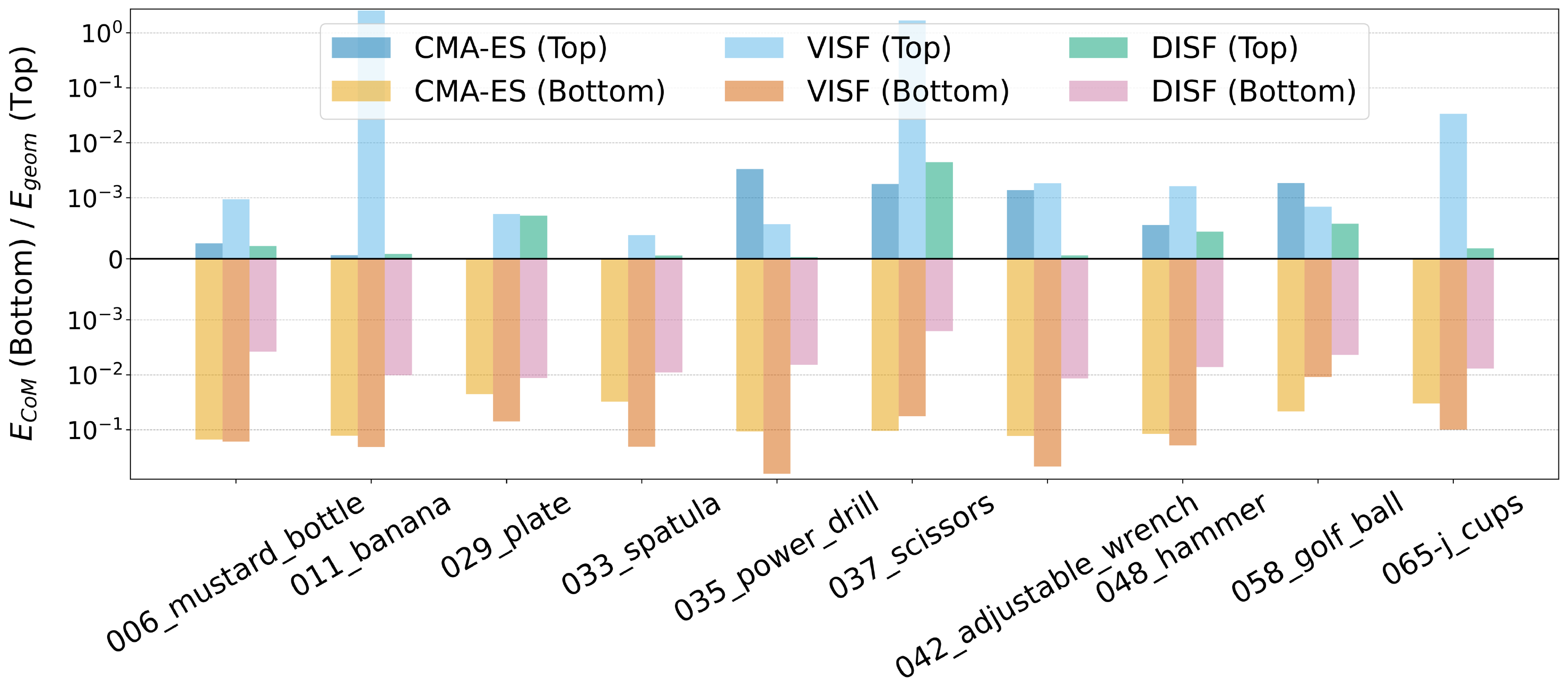}
                \caption{   
                   The results of planned grasp quality. 
                   The Top part represents the geometric compatibility error, while the Bottom part represents the CoM alignment error. 
                   In both the Top and Bottom plots, lower values indicate better performance.
                   For each object, the Top and Bottom correspond to the same method. 
                }
                \label{fig:results_grasp_quality}
            \end{figure}

        \subsubsection{Results}
            The results are shown in Fig.~\ref{fig:results_grasp_quality}. 
            The Top part represents the geometric compatibility error, while the Bottom part represents the CoM alignment error. For geometric compatibility evaluation, the sampling-based approach (CMA-ES) and the proposed DISF exhibit lower errors, whereas the conventional VISF consistently shows higher errors. An interesting phenomenon is that VISF, which is supposed to optimize geometric compatibility, performs worse than the other two methods. 
            This can be attributed to the difficulty of simultaneously optimizing multiple pose parameters within VISF's optimization framework. Specifically, VISF optimizes both rotational and translational parameters under the geometric compatibility metric. 
            However, due to kinematic constraints that prevent independent movement of different gripper surfaces, it tends to converge to poor local minima. 
            Another noteworthy aspect of the Bottom (CoM alignment error) is its relation to the Top (geometric compatibility error). While CMA-ES and VISF show a significant difference in geometric compatibility error, their CoM alignment errors are not markedly different. Conversely, CMA-ES and DISF exhibit similar geometric compatibility errors, yet our proposed DISF consistently achieves lower CoM alignment errors. 
            
            Overall, while the proposed DISF maintains good performance in geometric compatibility, it significantly improves CoM alignment, potentially leading to a more stable distribution of contact points.

    \subsection{Grasp Success Rate Evaluation}
        \subsubsection{Settings}   
            To evaluate the feasibility of the planned grasps, we performed grasp executions in the MuJoCo physics simulator~\cite{mujoco,menagerie2022github} and assessed the grasp success rate for each object.

            The execution process consists of the following steps.  
            First, the robot hand's palm is moved to the pre-grasp pose, which is computed based on the optimal grasp pose $(\mR(\vomega^*), \vt^*)$ for each object.  
            Next, from the pre-grasp pose, the hand moves to the final grasp pose and adjusts the gripper aperture with $\dd^*$. 
            Following this, the object is lifted vertically upward by $30$ [cm] and held stationary for $1$ second.  
            Finally, the success of the grasp is evaluated based on the positional and orientational errors of the lifted object.  
            These errors are defined as follows:  
            \begin{subequations}
                \label{eq:grasp_evaluation_error}
                \begin{align}
                    \mathbf{e}_{\textrm{pos}} &= \|\vp_{\textrm{target}} - \vp_{\textrm{lift}} \|, \\
                    \mathbf{e}_{\textrm{ori}} &= 2 \cdot \arccos(|w|), 
                \end{align}
            \end{subequations}
            where $\mathbf{e}_{\textrm{pos}}$ represents the positional error, and $\mathbf{e}_{\textrm{ori}}$ represents the orientational error.  
            The target position $\vp_{\textrm{target}}$ is determined based on the initial object position and lifting height, while $\vp_{\textrm{lift}}$ is the observed object position after the grasp execution.  
            The term $w$ is the scalar component of the quaternion representing the object's orientation.
            
            A grasp is then considered successful if both errors remain below predefined thresholds:  
            \begin{equation}
            \label{eq:grasp_evaluation_condition}
                \text{Success} = 
                \begin{cases} 
                    1, & \text{if } (\mathbf{e}_{\text{pos}} < \eta_{\text{pos}}) \land (\mathbf{e}_{\text{ori}} < \eta_{\text{ori}}), \\
                    0, & \text{otherwise}.
                \end{cases}
            \end{equation}
            where $\eta_{\text{pos}}$ and $\eta_{\text{ori}}$ are the predefined thresholds for positional and orientational errors, respectively.
            The supplementary information can be found in our \href{https://tomoya-yamanokuchi.github.io/disf-project-page/}{project page}.

        \subsubsection{Results}        
            Table~\ref{tab:grasp_success_results} summarizes the grasp execution results.  
            It can be observed that the proposed DISF method outperforms the baseline methods, highlighting the importance of CoM alignment in ensuring stable distribution of contact points for successful grasps. 

\begin{figure}[H]
                \centering
                \includegraphics[width=\columnwidth]{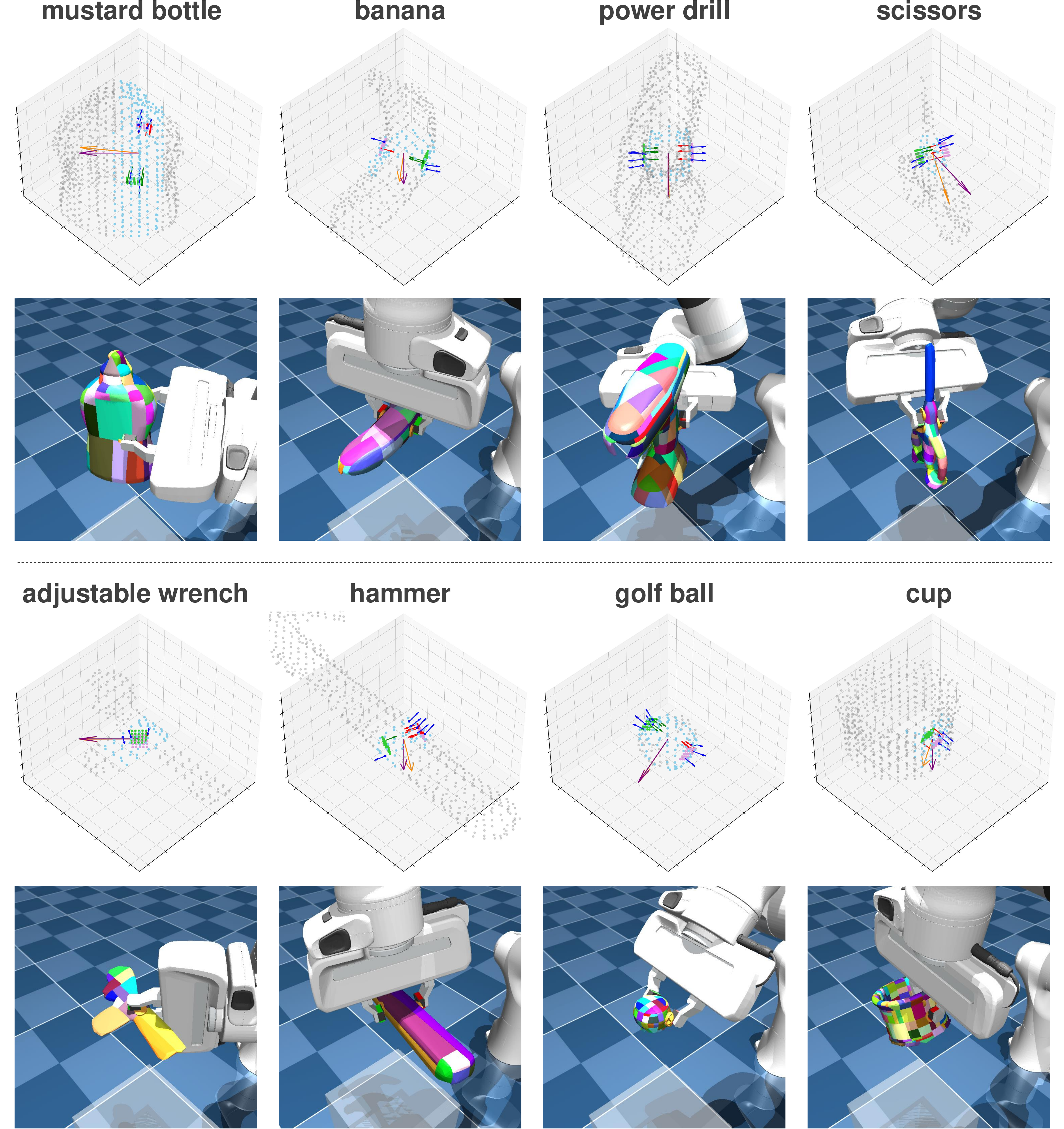}
                \caption{   
                    This figure showcases successful grasp executions using the proposed DISF method. It presents grasp planning in point cloud space alongside simulation results. Key visual elements include contact points (cyan), surface fitting points (blue), and fingertip surfaces (plum and lime green). Normal vectors and hand axes are represented with arrows. The results confirm that DISF effectively optimizes grasp quality and achieves successful object grasping in simulation.
                }
                \label{fig:grasp_results_success_disf}
            \end{figure}

            Fig.~\ref{fig:grasp_results_success_disf} illustrates the results of eight successful grasp cases planned by DISF. 
            It demonstrates that DISF effectively plans feasible grasps that maximize grasp quality in terms of geometric compatibility and contact stability. 
            
\begin{figure}[H]
                \centering
                \includegraphics[width=\columnwidth]{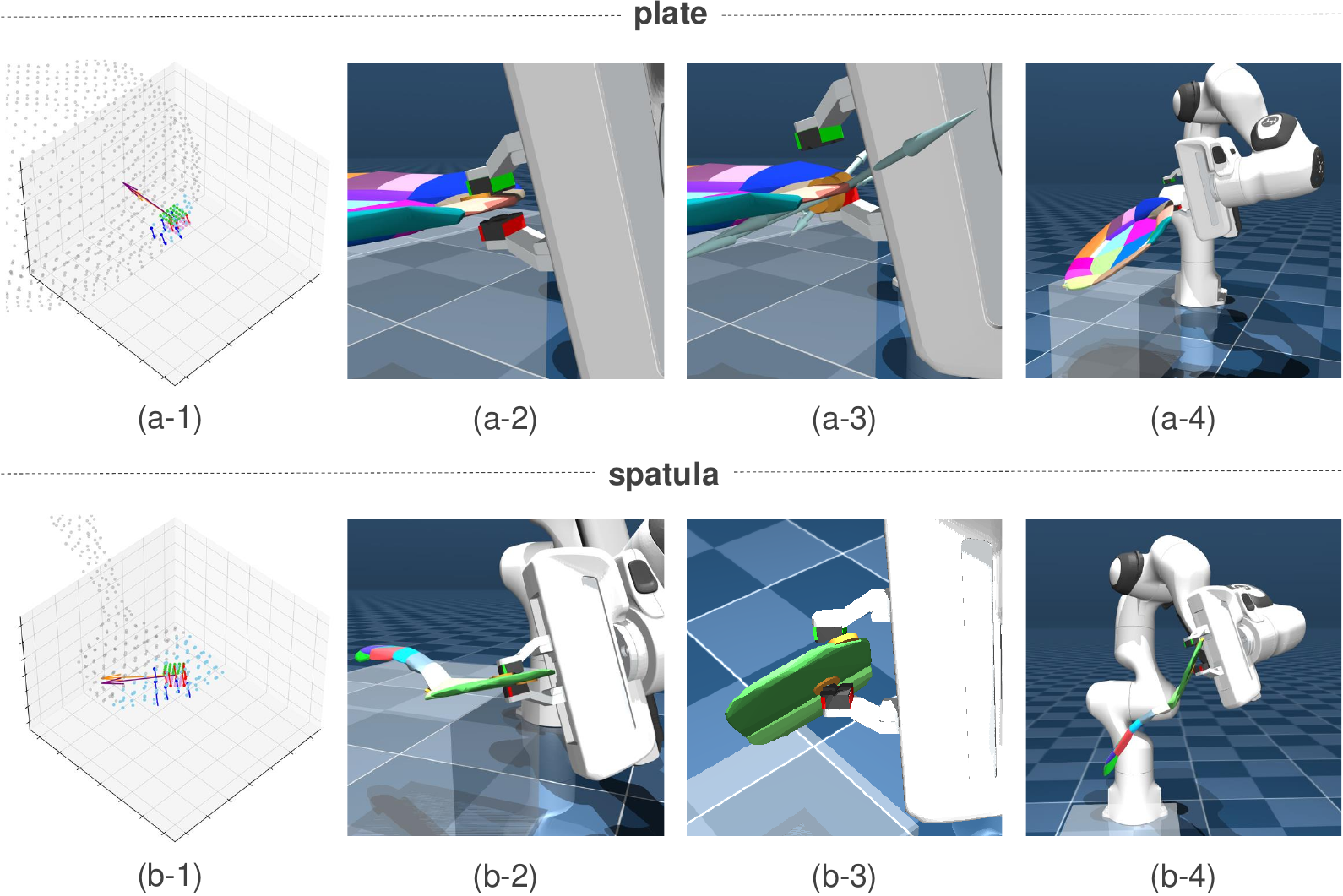}
                \caption{
                    This figure illustrates failed grasp attempts using the DISF method. The top row represents the $\texttt{029\_plate}$ case, while the bottom row corresponds to the $\texttt{033\_spatula}$ case. In each row, the leftmost image shows grasp planning, followed by three images depicting execution in simulation. Failure in the plate case occurs due to the right finger missing contact, causing the object to slip. In the spatula case, initial contact is successful, but tilting during lifting induces an unstable grasp due to its rotational moment.
                }
                \label{fig:grasp_results_failure_disf}
            \end{figure}

            Conversely, Fig.~\ref{fig:grasp_results_failure_disf} presents two failed grasp cases.  
            Although the planned grasps appear feasible at first glance, execution in simulation reveals failure due to two key limitations inherent to the surface fitting framework:  
            (1) the lack of precise contact modeling, and  
            (2) an inability to account for the true CoM of the entire object surface.  
            Because surface fitting is fundamentally based on aligning point cloud data, it does not explicitly model contact forces or frictional constraints, which are crucial for grasping.  
            Additionally, since the contact surface used in the surface fitting algorithm differs from the full object surface, it does not fully capture the object's true CoM.  
            As a result, when the planned grasp position is far from the actual CoM, a large rotational moment is induced, leading to grasp instability.  
            In the $\texttt{029\_plate}$ case, the grasp is optimized in the point cloud space but fails due to incomplete surface contact.  
            In the $\texttt{033\_spatula}$ case, a large rotational moment is induced, resulting in grasp destabilization during lifting.

            \begin{table}[H]
                \centering
                \caption{
                    This table presents the success rates of grasp executions in the simulation experiments. 
                    The checkmark ($\greencheck$) indicates success, while the horizontal bar (-) indicates failure. 
                    The "Correspondence" column shows the total number of correspondence pairs used for surface fitting optimization. 
                    The bottom row reports the overall grasp success rate for each method across all objects and its average planning time. 
                }
                \label{tab:grasp_success_results}
                \begin{tabular}{l@{\hspace{1.5em}}c@{\hspace{1.5em}}c@{\hspace{1.5em}}c@{\hspace{1.5em}}c@{\hspace{1.5em}}}
                    \toprule
                    Object & CMA-ES & VISF & DISF (ours) & Correspondence\\
                    \midrule
                    $\texttt{006\_mustard\_bottle}$     & - & - & $\greencheck$ & $8$ \\
                    $\texttt{011\_banana}$              & - & - & $\greencheck$ & $3$ \\
                    $\texttt{029\_plate}$               & - & - & - & $5$ \\
                    $\texttt{033\_spatula}$             & - & - & - & $6$ \\
                    $\texttt{035\_power\_drill}$        & - & - & $\greencheck$ & $7$ \\
                    $\texttt{037\_scissors}$            & - & - & $\greencheck$ & $6$ \\
                    $\texttt{042\_adjustable\_wrench}$  & - & - & $\greencheck$ & $3$ \\
                    $\texttt{048\_hammer}$              & - & - & $\greencheck$ & $6$ \\
                    $\texttt{058\_golf\_ball}$          & - & - & $\greencheck$ & $7$ \\
                    $\texttt{065-j\_cups}$              & $\greencheck$ & - & $\greencheck$ & $3$ \\
                    \midrule
                    Success Rate        & $1/10$    & $0/10$ & {$\textbf{8/10}$} &  -  \\
                    \midrule
                    Planning Time [ms] & $172.4$    & $\textbf{1.4}$ & $4$ &  -  \\
                    \bottomrule
                \end{tabular}
            \end{table}

\section{Discussion}
    Here, we discuss the limitations of our method.
    The first limitation lies in the extraction of an appropriate contact surface and correspondence pairs.  
    In our experiments, we manually extracted the object's contact surface and compute correspondences. 
    As a result, our method does not generalize to unknown objects without prior adjustment. 
    A promising direction for future research is to integrate a higher-level planner that can autonomously identify the appropriate contact surface and establish correspondences.  
    One possible approach is to leverage language instructions~\cite{language_CoRL_2023}. 
    Another approach is to utilize affordance-based reasoning~\cite{affordances_IROS_2017}, where functional attributes of objects guide the selection of suitable grasping regions.  
    Incorporating such methods could enhance the adaptability of our approach to novel objects in unstructured environments.

    The second limitation is the limited scope of our experimental evaluation.  
    Although we validated our method on ten YCB objects, this sample size is insufficient to fully assess its generalizability, necessitating further testing with larger datasets.  
    Additionally, our approach assumes access to complete point cloud data, whereas real-world object perception often relies on partial point clouds from RGB-D cameras~\cite{Fan_Case2018,Fan_RAL_2019,Fan_Sensors_2024}, adding complexity to grasp planning. 
    Thus, evaluating our method in real robotic environments with incomplete object shape information remains an important direction for future work.

\section{Conclusion}
    In this paper, we proposed a novel surface fitting-based grasp planning algorithm that extends conventional geometric compatibility optimization by incorporating CoM alignment to ensure that the gripper and object surfaces are spatially aligned, thereby enhancing contact stability.  
    Inspired by human grasping behavior, our method disentangles the grasp pose optimization process into three sequential steps:  
    (1) rotation optimization to align contact normals,  
    (2) translation refinement for CoM alignment, and  
    (3) gripper aperture adjustment to optimize contact point distribution.  
    
    To validate the effectiveness of our proposed method, we conducted grasp planning experiments on ten YCB objects, evaluating grasp quality in terms of geometric compatibility and contact stability through CoM alignment.  
    Additionally, we assessed the feasibility of the planned grasps by measuring the grasp success rate in a physics simulation.  
    The results show 80\% improvement in grasp robustness over conventional surface fitting approaches, highlighting the importance of CoM alignment for stable grasps.

\section*{Acknowledgments}
    This research is funded by JSPS KAKENHI Grant Number 23KJ1584.

%
%
\bibliographystyle{IEEEtran}

\end{document}